\documentclass[conference]{IEEEtran}

\usepackage[cmex10]{amsmath}
\usepackage{amsthm}
\usepackage{amssymb}
\usepackage{mathrsfs}
\usepackage{graphicx}
\usepackage{float}
\usepackage{array}
\usepackage{epstopdf}
\usepackage{multirow}
\usepackage{cite}

\begin{document}

\title{FingerNet: Pushing The Limits of Fingerprint Recognition Using Convolutional Neural Network}

\author{Shervin Minaee$^*$, Elham Azimi$^*$, Amirali Abdolrashidi$^{\dagger}$  \\
$^*$New York University
\\ $^{\dagger}$University of California, Riverside\\ \\
}

\maketitle

\begin{abstract}
Fingerprint recognition has been utilized for cellphone authentication, airport security and beyond. 
Many different features and algorithms have been proposed to improve fingerprint recognition.
In this paper, we propose an end-to-end deep learning framework for fingerprint recognition using convolutional neural networks (CNNs) which can jointly learn the feature representation and perform recognition.
We train our model on a large-scale fingerprint recognition dataset, and improve over previous approaches in terms of accuracy. Our proposed model is able to achieve a very high recognition accuracy on a well-known fingerprint dataset. We believe this framework can be widely used for biometrics recognition tasks, making more scalable and accurate systems possible.
We have also used a visualization technique to highlight the important areas in an input fingerprint image, that mostly impact the recognition results. 

\end{abstract}

\IEEEpeerreviewmaketitle

\section{Introduction}
To  make  an  application  more  secure  and  less  accessible  to undesired people, we need to be able to distinguish a person from the others.  
There are various ways to identify a person, and biometrics have been one of the most secure options so far.  
They are virtually impossible to imitate by anyone other than the desired person. 
They can be divided into two categories:  \textit{behavioral features}, which are actions that a person can uniquely create or express, such as signature and walking rhythm; and \textit{physiological features}, which are characteristics that a person possesses, such as fingerprint and iris pattern. Many works revolve around recognition and categorization of such data including, but not limited to, fingerprints, faces, palmprints and iris patterns \cite{biomet1,biomet2,biomet3,biomet4,biomet5}.

Fingerprint has been used in various applications such as forensics, transaction authentication, cellphone unlocking, etc. Many algorithms proposed for fingerprint recognition are minutiae-based matching. The major minutiae features of fingerprint ridges are ridge ending, bifurcation, and short ridge. 
There have also been various works on fingerprint recognition using hand-crafted features followed by some classification in the past.
In \cite{park}, Park proposed a fingerprint recognition system based
on SIFT features. In \cite{cappelli}, Cappelli proposed a new representation based on a 3D data structure built from minutiae
distances and angles called Minutiae Cylinder-Code (MCC).
More recently, Minaee et al \cite{scat} proposed a fingerprint recognition using multi-layer scattering convolutional networks, which decomposes fingerprint images using wavelets of different scales and orientations. 
\begin{figure}[h]
\begin{center}
    \includegraphics [scale=0.66] {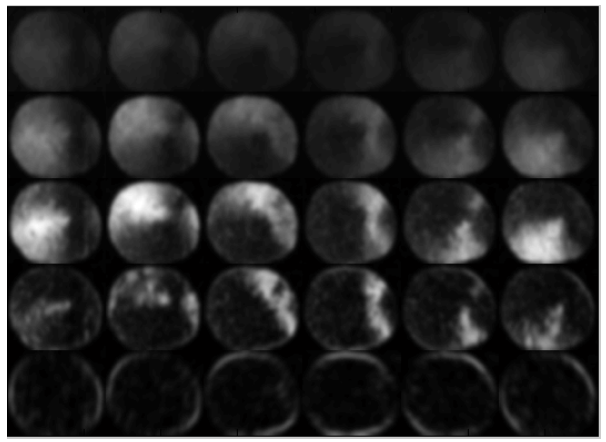}
    \includegraphics [scale=0.4] {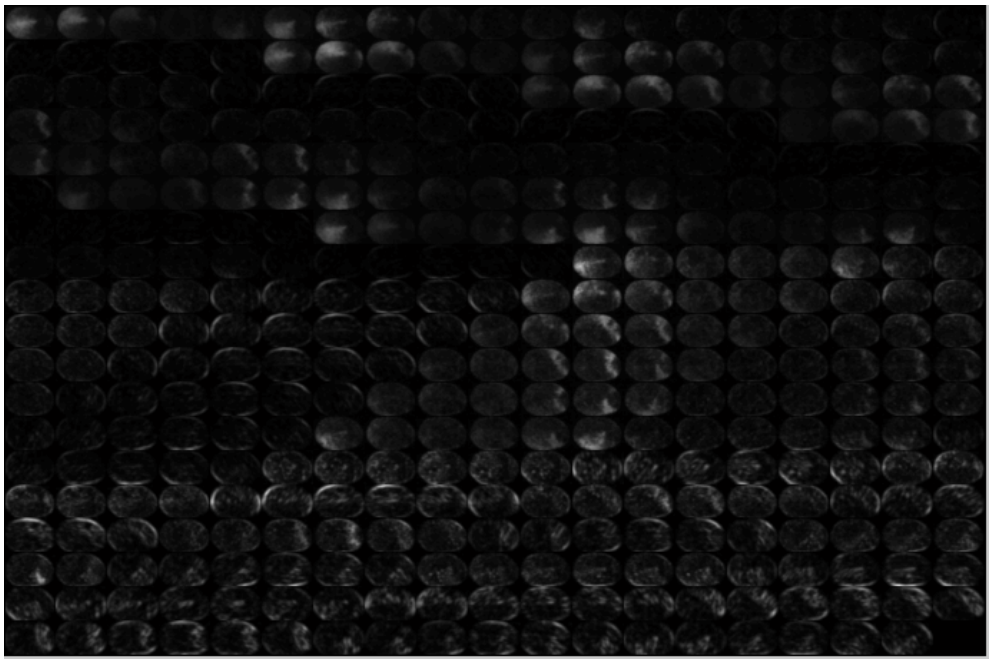}
\end{center}
  \caption{The images from the first (top) and second layers  (bottom) of scattering transform  \cite{scat}, each image capturing the wavelet energies along specific orientation and scale.}
\end{figure}

Although many of the previous works achieve highly accurate performances, they involve a lot of pre-processing and using several hand-crafted features, which may not be optimal for different fingerprint datasets (collected under different conditions).
In recent years, there have been a lot of focus on developing models for jointly learning the features, while doing prediction. 
Along this direction, convolutional neural networks (CNNs) \cite{cnn} have been very successful in various computer vision and natural language processing (NLP) tasks in recent years \cite{alexnet}.
Their success is mainly due to three factors: the availability of large-scale manually labeled datasets; powerful processing tools (such as GPGPUs); and good regularization techniques (such as dropout, etc.) that can prevent the overfitting problem.

Deep learning have been used for various problems such as  classification,  segmentation, super-resolution, image captioning, emotion analysis, face recognition, and object detection,  and significantly improved the performance over traditional approaches \cite{cnn1,cnn2,cnn3,cnn4,cnn5,cnn6,cnn7,cnn8,cnn9}. 
It has also been used heavily for various NLP tasks, such as sentiment analysis, machine translation, name-entity-recognition, and question answering \cite{nlp1,nlp2,nlp3,nlp4}.
More interestingly, it is shown that the features learned from some of these deep architectures can be transferred to other tasks easily, i.e. one can get the features from a trained model for a specific task and use it for a different task, by training a classifier/predictor on top of it \cite{offshelf}.

However, having a large-scale dataset (with several examples for each class of label) is crucial for the success of most of the current deep learning-based models. For fingerprint recognition, there are several public
datasets with a reasonable size, but most of them come with a limited number of images per class (usually less than 20 fingerprint images per person), which makes it more challenging to train a convolutional neural network from scratch on these datasets. 

In this work, we propose a deep learning framework for fingerprint recognition for the case where only a few samples are available for each class (few shots learning). It can get the fingerprint images and perform recognition directly. 
Previously there have been works using the features extracted from a pre-trained convolutional network \cite{cnn_bio}, and used along with various classifiers (such as SVM) to perform biometrics recognition, but in this work, we train a model for fingerprint recognition directly.

We train a convolutional neural network on one of the popular fingerprint datasets. We initialize the model's weights with the ones trained on ImageNet dataset, and fine-tune it on our dataset. 
We also employ data augmentation techniques (such as flipping, random cropping, adding small amount of distortions) to increase the number of samples for each class. By doing so, we are able to achieve a very high recognition accuracy rate, on the test samples of this dataset. 
We believe this framework can be widely used for any other biometric recognition task.

The structure of the rest of this paper is as follows.
Section II provides the the description of the overall proposed framework. Section III  provides the experimental studies and comparison with previous works.
And finally the paper is concluded in Section IV.

\section{The Proposed Framework}
In this work, we propose a transfer learning approach by fine-tuning a pre-trained convolutional neural network on a popular fingerprint recognition dataset. 
We will first provide a quick introduction of transfer learning, and then discuss the proposed framework. 

\subsection{Transfer Learning Approach}
In transfer learning, a model trained on one task is re-purposed on another related task, usually by some adaptation toward the new task.
For example, one can imagine using an image classification model trained on ImageNet \cite{imagenet} to perform texture classification. 
It would be plausible to use the representation learned by a model, trained for general-purpose classification, for a different image processing task. 
There have been many works based on pre-trained deep learning models to perform a different task in the past few years. 

There are two main ways in which the pre-trained model is used for a different task. 
In one approach, the pre-trained model, e.g. a language model, is treated as a feature extractor, and a classifier is trained on top of it to perform classification, e.g. sentiment analysis. Here the internal weights of the pre-trained model are not adapted to the new task. 

In the other approach, the whole network, or a subset thereof, is fine-tuned on the new task. Therefore the pre-trained model weights are treated as the initial values for the new task, and are updated during the training stage.

\subsection{Fingerprint Image Classification Using Residual ConvNet}
\begin{figure}[h]
\begin{center}
   \includegraphics[width=0.7\linewidth]{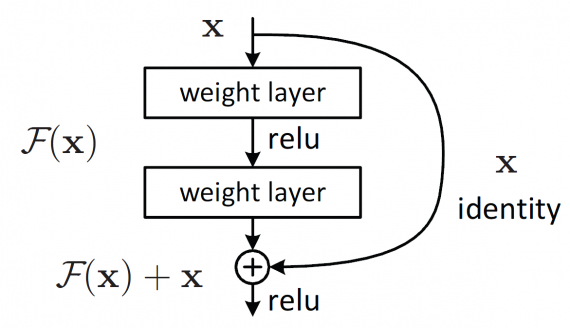}
\end{center}
   \caption{The residual block used in ResNet model}
\label{fig:Resnet}
\end{figure}

In this work, we focused on fingerprint recognition task, and chose a dataset with a large number of subjects, but limited number of images per subject, and proposed a transfer learning approach to perform identity recognition using a deep residual convolutional network.
We use a pre-trained ResNet50 \cite{cnn1} model trained on ImageNet dataset, and fine-tune it on our own training images. 
ResNet is popular CNN architecture which provides easier gradient flow for more efficient training, and was the winner of the 2015 ImageNet competition. 
The core idea of ResNet is introducing a so-called \textit{identity shortcut connection} that skips one or more layers, as shown in Figure \ref{fig:Resnet}.
This would help the network to provide a direct path to the very early layers in the network, making the gradient updates for those layers much easier.

\begin{figure*}[h]
\begin{center}
   \includegraphics[page=1,width=1.0\linewidth]{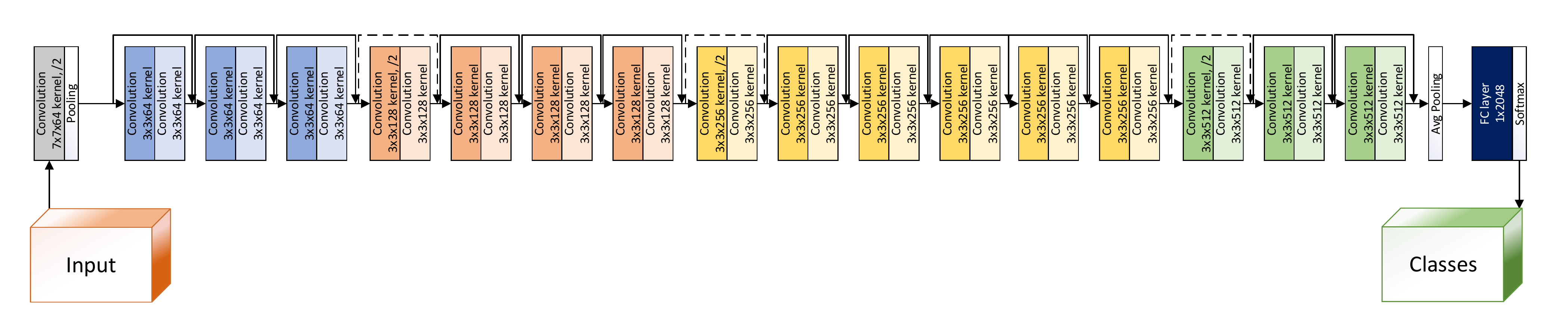}
\end{center}
   \caption{The architecture of ResNet50 neural network \cite{cnn1}, and how it is transferred for fingerprint recognition. The last layer is changed to match the number of classes in our dataset.}
\label{fig:ResnetNN}
\end{figure*}

To perform recognition on our fingerprint dataset, we fine-tuned a ResNet model with 50 layers on the augmented training set.
The overall block diagram of ResNet50 model, and how it is used for fingerprint recognition is illustrated in Figure \ref{fig:ResnetNN}.
We applied several data augmentation techniques to increase the number of training samples, including horizontal and vertical flip, random crops, and small distortions. 
By adding these augmentation, the number of training samples is increased by a factor of $\sim$3x.
We then fine-tune this model for a fixed number of epochs, which is determined based on the performance on a validation set, and then evaluate on the test set.
This model is then trained with a cross-entropy loss function. To reduce the chance of over-fitting the $\ell_2$ norm can be added to the loss function, resulting in an overall loss function as: 
\begin{equation}
\begin{aligned}
& \mathcal{L}_{final}=  \mathcal{L}_{class}+  \lambda_1 ||\textit{W}_{fc}||_F^2
\end{aligned}
\end{equation}
where $\mathcal{L}_{class}= - \sum_{i} p_i \log(q_i)$ is the cross-entropy loss, and $||W_{fc}||_F^2$ denotes the Frobenius norm of the weight matrix in the last layer.
We can then minimize this loss function using stochastic gradient descent kind of algorithms.

\section{Experimental Results}
In this section we provide the experimental results for the proposed algorithm, and the comparison with the previous works on this dataset.

We train the proposed model for 100 epochs using a Nvidia Tesla GPU.
The batch size is set to 24, and Adam optimizer is used to optimize the loss function, with a learning rate of 0.0001.
All images are down-sampled to 224x224 before being fed to the neural network.
All our implementations are done in PyTorch \cite{pytorch}.
We present the details of the dataset used for our work in the next section, followed by  quantitative and visual  experimental results.

\subsection{Dataset}
\begin{figure}[h]
\begin{center}
    \includegraphics [scale=0.5] {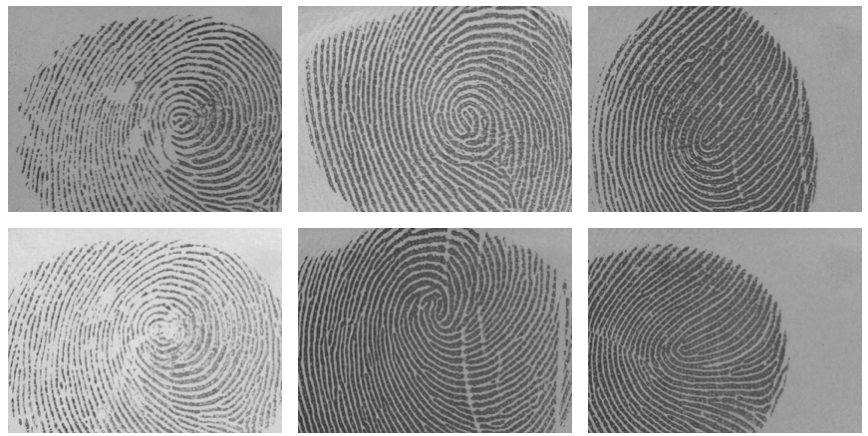}
\end{center}
  \caption{Six sample fingerprint images from PolyU dataset.}
  \label{fig:PolyU}
\end{figure}

We evaluated our work using the PolyU fingerprint
database which is provided by Hong Kong Polytechnic University \cite{polyu1,polyu2}. It contains 1480 images of 148 fingers. 
Six sample images from this dataset are shown in Figure \ref{fig:PolyU}. It can be seen that the fingerprint images in this dataset have slightly different color distributions, as well as different sizes.

For each person, 4 images are used as test samples randomly, and the rest are used for training and validation.

\subsection{Comparison with previous works}
Table I provides a comparison of the performance of the proposed model and some of the previous works on this dataset.
As we can see, the model outperforms previous works on this dataset. 
It is worth mentioning that the model in \cite{scat} also uses features derived from a convolutional neural network with pre-defined filters. The scattering convolutional network can be thought of as a feature extractor trying to capture higher-order statistics.
\begin{table}[ht]
\centering
  \caption{Comparison of performance of different algorithms on PolyU database}
  \centering
\begin{tabular}{|m{5cm}|m{1.5cm}|}
\hline
Method  & Accuracy Rate\\
\hline
Scattering network \cite{scat} &   \ \ \ \ \ \ 92\% \\
\hline
Gabor-wavelet \cite{gabor} (on a subset of 50 subjects) &   \ \ \ \ \ \ 95.5\% \\
\hline
 \textbf{The proposed algorithm}  &  \ \ \ \ \ \ \textbf{95.7\%}\\
\hline
\end{tabular}
\label{TblComp}
\end{table}

\subsection{Important Regions Visualization} 
\begin{figure}[h]
\begin{center}
    \includegraphics [scale=0.4] {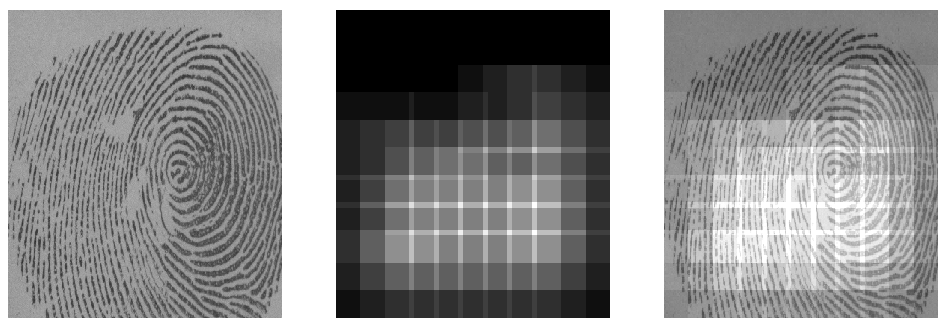}
    \includegraphics [scale=0.4] {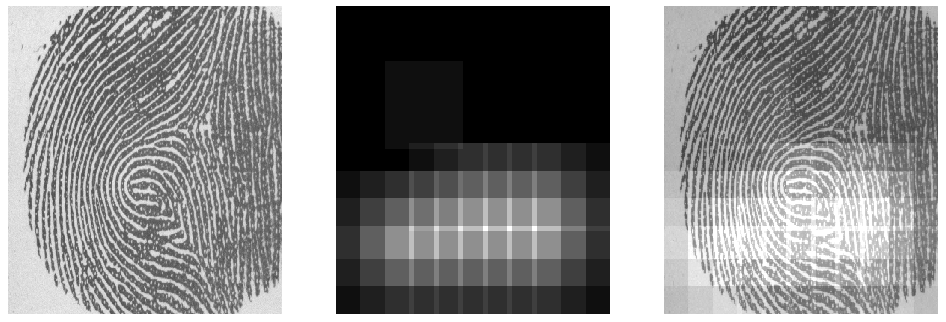}
    \includegraphics [scale=0.4] {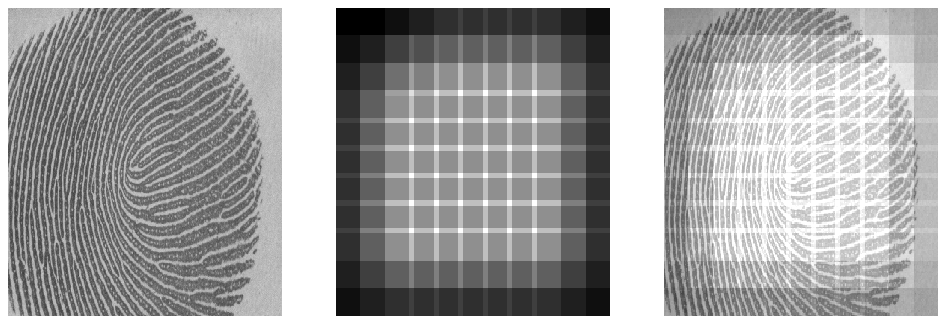}
    \includegraphics [scale=0.4] {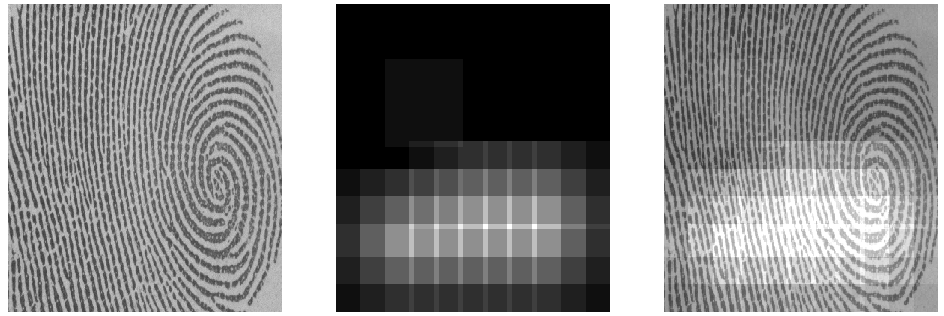}
\end{center}
  \caption{The saliency map of important regions for fingerprint recognition.}
\label{fig:finger_visual}
\end{figure}

Here we use a simple approach to detect the most important regions while performing fingerprint recognition using convolutional network, originally inspired by \cite{fergus}. 
We start from the top-left corner of the image, and each time zero out a square region of size $N$x$N$ inside the image, and make a prediction using the trained model on the occluded image.
If occluding that region causes the model to mis-classify that fingerprint image, that area would be considered as an important region while doing fingerprint recognition.
On the other hand, if removing that region does not impact the model's prediction, we infer that region is not as important.
Now if we repeat this procedure for different sliding windows of $N$x$N$, each time shifting them with a stride of $S$, we can get a saliency map for the most important regions in recognizing fingerprints.
The saliency maps for four example fingerprint images are shown in Figure \ref{fig:finger_visual}.

\section{Conclusion}
In this work we propose a deep learning framework for fingerprint recognition, by fine-tuning a pre-trained convolutional model on ImageNet. 
This framework is applicable for other biometrics recognition problems, and is especially useful for the cases where there are only a few labeled images available for each class.
We apply the proposed framework on a well-known fingerprint dataset, PolyU, and achieved promising results, which outperforms previous approaches on this dataset.
We train these models with very few original images per class.
We also utilize a visualization technique to detect and highlight the most important regions of a fingerprint image during fingerprint recognition.

\section*{Acknowledgment}
The authors would like to thank the biometric research group at PolyU Hong Kong for providing the fingerprint dataset used in this work. We would also like to thank Facebook AI research for open sourcing the PyTorch package.

\end{document}